\pdfoutput=1 
\documentclass[AMA,LATO1COL]{WileyNJD-v2}
\usepackage{moreverb}
\usepackage{float}
\usepackage{algorithmicx}
\usepackage{algorithm}
\usepackage{algpseudocode}
\usepackage{amsthm}

\newtheorem{lemm}{Lemma}

\algnewcommand\algorithmicinput{\textbf{INPUT:}}
\algnewcommand\INPUT{\item[\algorithmicinput]}

\algnewcommand\algorithmicinputT{\textbf{DESCRIPTION:}}
\algnewcommand\DESCRIPTION{\item[\algorithmicinputT]}

\newcommand\BibTeX{{\rmfamily B\kern-.05em \textsc{i\kern-.025em b}\kern-.08em
T\kern-.1667em\lower.7ex\hbox{E}\kern-.125emX}}

\articletype{SPECIAL ISSUE PAPER}%
\received{}
\revised{}
\accepted{}
        
\begin{document}

\title{Crossover-SGD: A gossip-based communication in distributed deep learning for alleviating large mini-batch problem and enhancing scalability}

\author[1,2]{Sangho Yeo}
\author[3]{Minho Bae}
\author[4]{Minjoong Jeong}
\author[4]{Oh-Kyoung Kwon}
\author[1]{Sangyoon Oh*}

\authormark{YEO \textsc{ET AL.}}

\address[1]{\orgdiv{Department of Artificial Intelligence}, \orgname{Ajou University}, \orgaddress{\state{Suwon}, \country{Republic of Korea}}}
\address[2]{\orgdiv{ML Development Department}, \orgname{Surromind}, \orgaddress{\state{Seoul}, \country{Republic of Korea}}}
\address[3]{\orgdiv{Artificial Intelligence Convergence Research Section}, \orgname{ETRI}, \orgaddress{\state{Seongnam}, \country{Republic of Korea}}}
\address[4]{\orgdiv{Supercomputing Department}, \orgname{Korea Institute of Science and Technology}, \orgaddress{\state{Daejeon}, \country{Republic of Korea}}}

\corres{*Sangyoon Oh,  World cup-ro 206, Yeongtong-gu, Suwon-si, Gyeonggi-do, Republic of Korea.\email{syoh@ajou.ac.kr}}
\presentaddress{National Supercomputing Center with supercomputing resources including technical support (KSC-2019-CRE-0105), the MSIT (Ministry of Science and ICT),Korea, under the ITRC (Information Technology Research Center) support program (IITP-2020-2018-0-01431) supervised by the IITP (Institute for Information communications Technology Promotion), Basic Science Research Program Through the National Research Foundation of Korea (NRF) funded by the Ministry of Education (2018R1D1A1B07043858), and Electronics and Telecommunications Research Institute (ETRI) grant funded by the Korean government (20ZT1100, Development of ICT Convergence Technology based on Urban Area),}

\abstract[Abstract]{Distributed deep learning is an effective way to reduce the training time for large datasets as well as complex models. However, the limited scalability caused by network-overheads makes it difficult to synchronize the parameters of all workers and gossip-based methods that demonstrate stable scalability regardless of the number of workers have been proposed. However, to use gossip-based methods in general cases, the validation accuracy for a large mini-batch needs to be verified. For this, we first empirically study the characteristics of gossip methods in a large mini-batch problem and observe that gossip methods preserve higher validation accuracy than AllReduce-SGD (Stochastic Gradient Descent) when the number of batch sizes is increased, and the number of workers is fixed. However, the delayed parameter propagation of the gossip-based models decreases validation accuracy in large node scales. To cope with this problem, we propose Crossover-SGD that alleviates the delay propagation of weight parameters via segment-wise communication and random network topology with fair peer selection. We also adapt hierarchical communication to limit the number of workers in gossip-based communication methods. To validate the effectiveness of our method, we conduct empirical experiments and observe that our Crossover-SGD shows higher node scalability than SGP (Stochastic Gradient Push).}



\maketitle

\section{Introduction}

Recently, there has been an increased application of machine learning and deep learning in knowledge discovery and intelligent services. The main reason for this development is the successful convergence of the advances in deep learning algorithms and techniques (e.g., pretraining, ReLU activation function, and regularization) and considerable big data generated from various Internet of Things (IoT) devices and smartphones. However, there is still scope for improvement in deep learning performance because it still takes several days to train a single machine with large-scale data. Thus, researchers have attempted to reduce the training time of large-scale data by utilizing a distributed deep learning framework on the massive computing capacity of supercomputers or graphic processing unit (GPU) clusters. Using multiple workers, deep learning models that require several days to train on a single node can be trained in hours or minutes. For example, the ResNet50 model took 8,831.3 min (i.e., 6 days) to achieve 75.9\% validation accuracy when only one P100 GPU was used. Utilizing multiple advanced GPUs for training, Yamazaki et al. reduced this training time to almost 1 min \cite{yamazaki2019yet}.

The major advantage of distributed deep learning is that the amount of computation on each node decreases with an increase in the number of nodes. However, distributed deep learning algorithms were originally designed to synchronize the model parameters on all nodes (i.e., AllReduce-SGD), and there are inevitable communication overheads for synchronization between large-scale participating nodes; thus, obtaining the full benefits of parallel processing is difficult owing to the network overhead.

The simplest way to resolve this is to sufficiently increase the amount of work on each node (i.e., increase the size of the mini-batch) to reduce the synchronization frequency. This solution addresses the problem well because the communication overhead becomes relatively smaller than the computation overhead when the synchronization frequency is reduced. However, this does not solve the large mini-batch problem \cite{keskar2016large}, wherein the model accuracy degrades as the size of the batch increases; moreover, we cannot significantly increase the global mini-batch size. Thus, the large mini-batch problem has become a key issue to be addressed for improving the performance of distributed deep learning. To alleviate the large mini-batch problem, many effective techniques have been proposed. In particular, LARS (Layer-wise Adaptive Rate Scaling) \cite{you2017large} has been developed; LARS is a widely used layer-wise learning rate scaling technique for stabilizing the training process of distributed deep learning. This layer-wise learning rate scaling adapts the learning rate of each layer by the ratio of the weight norm to the gradient norm. As a result, LARS does not allow the layers to diverge when a high learning rate is set. However, this approach still uses AllReduce-SGD and has not been validated in other communication methods.

There are two other approaches that adapt a network communication scheme for parameter synchronization. The first approach involves the segmentation of models into chunks of layers (i.e., segments). Using segment-wise communication \cite{sergeev2018horovod,sun2019optimizing,shi2020communication,jiang2020bacombo,hu2019decentralized}, the network bandwidth of each worker node can be used more efficiently than in layer-wise or model-wise communication. In particular, Horovod \cite{sergeev2018horovod} which is a well-known distributed deep learning communication library, utilizes a segment-wise communication scheme (i.e., tensor fusion) to reduce network overhead.

Studies following the second approach attempt to provide novel parameter synchronization approaches instead of the AllReduce-SGD scheme to alleviate network communication overhead increased due to the increased number of participating nodes. Gossip-based communication, which is a widely studied communication method, effectively reduces network overhead for synchronization \cite{daily2018gossipgrad,lian2017can,lian2018asynchronous,assran2019stochastic}. In the gossip-based method, it is not necessary for each worker process to be synchronized with all other processes. Instead, each worker process communicates with a few other worker processes that are selected using a predefined network topology at every training step. Thus, these methods show linear speed-up even when numerous worker processes are added, unlike AllReduce-SGD. However, in gossip-based methods, the propagation delay of the weight parameters results in an additional decrease in the validation accuracy with a large mini-batch problem. Thus, although the gossiping-SGD scheme has the advantage of low network overhead, we cannot increase the number of workers as desired. As a result, it is only advantageous in relatively small mini-batches in small clusters. 

In summary, increasing the size of the mini-batch has the advantage of scalability by reducing the synchronization frequency but still has heavy network traffic owing to the use of the AllReduce-SGD scheme. Conversely, the gossip-based SGD scheme reduces the network traffic itself, but the accuracy in the large mini-batch problem, which should be analyzed on large-scale nodes, has not been verified. Therefore, we studied the validation accuracy of gossip-based methods under the condition of large mini-batch problems. In this analysis, we verified that gossip-based SGD can achieve higher accuracy than AllReduce-SGD with fixed number of worker nodes and that expanding node scalability is crucial by alleviating delay propagation of parameters in order to fully utilize this characteristic of gossip-based SGD.

To alleviate the node scalability problem of gossip-based SGD, we propose a novel segment-wise communication scheme, called Crossover-SGD, based on the gossiping-SGD algorithm for high node scalability. Our proposed Crossover-SGD is inspired by the uniform crossover method \cite{syswerda1989uniform} of genetic algorithm; however, the detailed method of Crossover-SGD is different from that of genetic algorithm. For fast and direct propagation of the weight parameter, we section the model into segments and utilize the overlapped segment-wise communication and random network topology with fair peer selection. By overlapping segment-wise communication and random network topology with fair peer selection, Crossover-SGD achieves higher and more robust accuracy with increasing mini-batch size in large mini-batch problem conditions; it also has higher node scalability with regard to validation accuracy. Furthermore, we adapt the hierarchical communication method \cite{jia2018highly,kurth2017deep} to Crossover-SGD to limit the number of workers communicating using gossip-based communication.

The contributions of our study are as follows:
\begin{itemize}
    \item We propose a novel random network topology with fair peer selection and observe that this random network topology is better than directed exponential network topology, which is used in SGP. This result is valuable because the utilization of random network topology has been avoided in previous studies owing to its skewed communication.
    \item We analyze gossip-based distributed deep learning in large mini-batch problem conditions with state-of-the-art techniques (e.g., LARS). In this evaluation, we verify that the gossip-based methods can achieve higher validation accuracy than AllReduce-SGD with fixed number of worker nodes owing to their exploration characteristics caused by the slightly different weights of each worker process.
    \item We compare SGP, Crossover-SGD, and AllReduce-SGD in a high-performance GPU cluster environment with local area network. Through evaluations of this cluster environment, we obtain expensive results from many perspectives (i.e., increasing mini-batch size, worker size, and time efficiency) in large mini-batch problem conditions, which have not been evaluated in previous studies. As a result, we verify that Crossover-SGD and hierarchical Crossover-SGD are stronger contenders to AllReduce-SGD than SGP while preserving the time efficiency of gossip-based methods.
\end{itemize}

The remainder of this paper is organized as follows. In Section 2, we present related works on both gossip-based and segment-wise communication optimization of distributed deep learning. We represent our proposed method, Crossover-SGD, in Section 3 We then present our results and discussion in Section 4. Specifically, Section 4.2 presents the analysis of gossip-based methods in large mini-batch problem conditions. Section 4.3 comprises a comparison of node scalability and Section 4.4 presents a comparison on time efficiency. Finally, we provide concluding statements in Section 5.

\section{Related Works}

Recently, the optimization of network overhead in distributed deep learning has been a topic of considerable research interest. For our study, we explored two categories of research:  gossip-based distributed deep learning and segment-wise communication optimization in distributed deep learning.

\subsection{Gossip-based communication methods in distributed deep learning}

Common distributed deep learning synchronizes the model parameters of all nodes at each iteration. The problem with this kind of synchronization is the increase in the network overhead with the number of nodes. To resolve this problem, distributed deep learning with parameter servers communicates each node in an asynchronous manner \cite{dean2012large,zhang2015staleness,zhang2015deep}. Elastic Averaging SGD (EA-SGD) \cite{zhang2015deep}, which is one of asynchronous methods, is a communication method utilizing the exploration of weight parameters for training. In detail, EA-SGD does not synchronize all model parameters. Instead, it makes a slight difference between the global and local models to explore the averaged model parameters.
 
In centralized asynchronous communication methods, the network overhead is concentrated in a few parameter servers. Therefore, decentralized gossip-based methods have been developed to spread the concentrated network overhead to every worker node. DP-SGD \cite{lian2017can} is significant research on gossip-based distributed deep learning. This method shows that gossip-based communication methods can achieve similar accuracy as AllReduce-SGD when the number of worker nodes is limited to a small number. ADP-SGD \cite{lian2018asynchronous} is an asynchronous version of DP-SGD. It shows that asynchronous and decentralized distributed deep learning can achieve convergence accuracy similar to that of AllReduce-SGD.
 
Assran et al. \cite{assran2019stochastic} proposed the latest well-known method of decentralized deep learning, stochastic gradient push (SGP). They adopted the SGP for distributed deep learning. There are two main differences between the SGP method and the previous DP-SGD and ADP-SGD. The first difference is the push-sum mechanism. This mechanism can spread the parameter information faster than previous studies because it is not necessary for the sender and receiver of the information to be the same. The second difference is the time-varying graph for the network topology. Using a time-varying graph, the model information can be sent to multiple locations.

\begin{figure}[ht]
\centering
\includegraphics[width=0.6\linewidth]{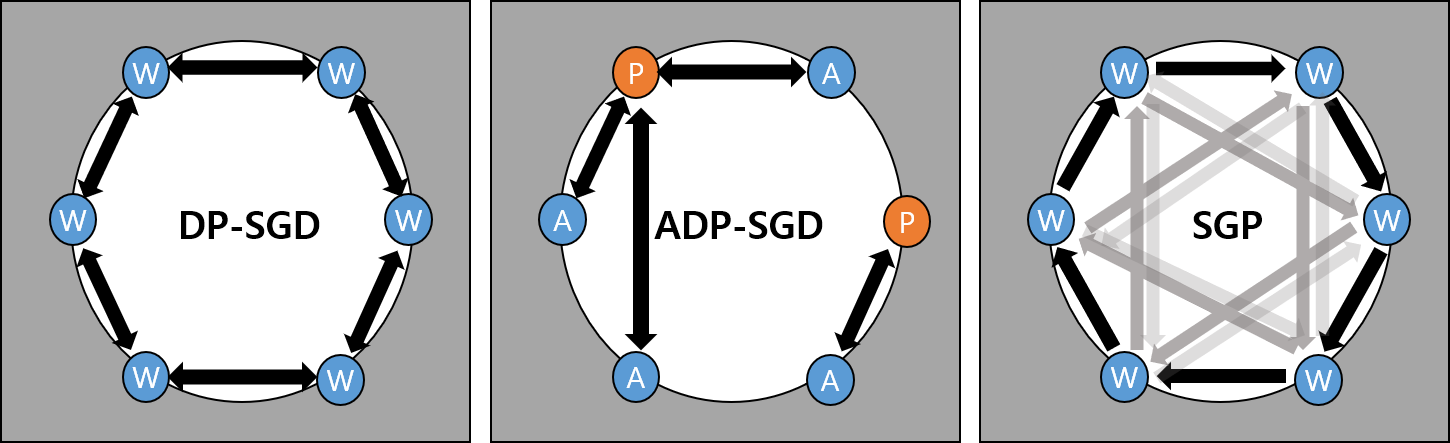}
\caption{Overview of previous gossip-based methods \label{gossip}}
\end{figure}

Figure~\ref{gossip} shows three representative gossip-based communications in distributed deep learning. These three methods are slightly different from each other. For example, DP-SGD communicates with others using a bidirectional ring network topology. In ADP-SGD, the workers are divided into two types. The first type is the active (i.e., A in the figure) workers, and the second type is the passive (i.e., P in the figure) workers. The passive workers act as parameter servers. Thus, it does not actively communicate with others; however, if a request for communication comes, it communicates with others. An active worker is one who can initiate the communication of passive workers. Two types of workers are configured to resolve the deadlock issue caused by asynchronous communication. Finally, SGP utilizes directional network topology, and unlike other methods, it uses time-varying network topology, as mentioned earlier.

Despite low network overhead caused by gossip-based methods, gossip communication methods are difficult to generally utilize in various domains due to reduced converged accuracy caused by delayed parameter propagation as the number of training processes is increased. For example, DP-SGD can be achieved similar validation accuracy with a small number of training processes, however, its accuracy is decreased as the number of training processes grows beyond that. Although the reduction in validation accuracy was mitigated, this trend also remains in the state-of-the-art studies, SGP.

\subsection{Segment-wise communication optimization in distributed deep learning}

\begin{figure}[ht]
\centering
\includegraphics[width=0.6\linewidth]{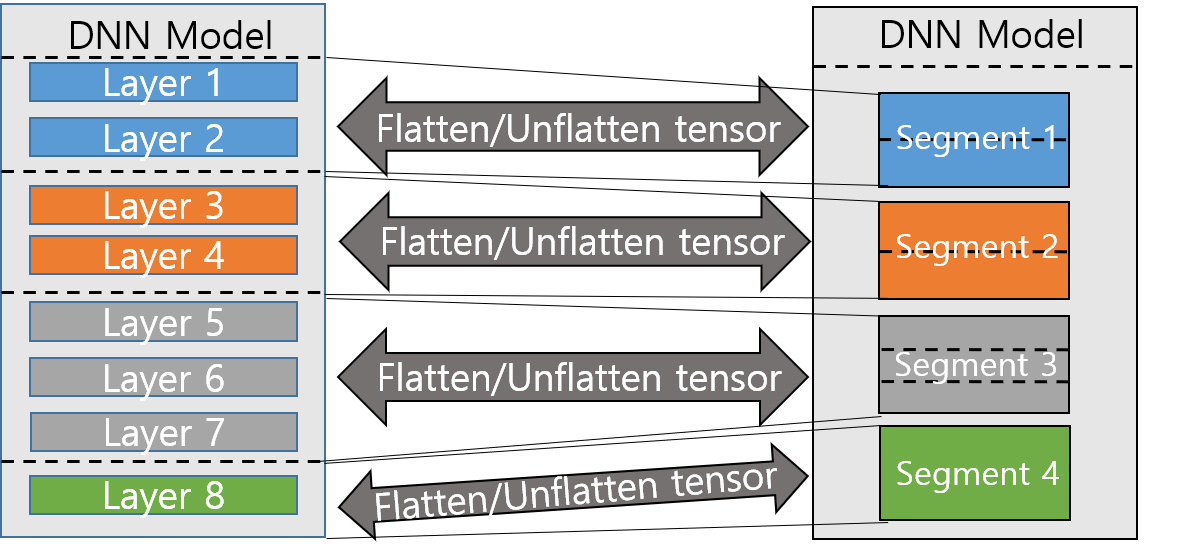}
\caption{Overview of DNN model segment \label{seg}}
\end{figure}

Segment-wise control for communication optimization has been continuously studied in distributed deep learning. It is difficult to communicate at once in the model used in distributed deep learning owing to its size; however, layer-wise communication leads to underutilization of network bandwidth. Thus, merging multiple layers into one large layer (i.e., tensor fusion \cite{sergeev2018horovod}) is efficient for communicating in a cluster environment. Figure \ref{seg} shows an overview of the tensor fusion. The transformation process between layers and segments is necessary because the number of layers is different from the number of segments, which is the chunk of layers created by tensor fusion.

Tensor fusion can be efficiently used in ring-AllReduce, which is a communication method for distributed deep learning that sets segments created by tensor fusion as the unit of communication. Many advanced methods have been proposed after tensor fusion. For example, lazy AllReduce \cite{sun2019optimizing} defines a memory pool to stack the tensor of layers onto the same places; subsequently, the size of the stacked tensor exceeds the threshold $\theta$, and the ring-AllReduce process starts. More recently, optimal merged gradient sparsification with SGD (OMGS-SGD) \cite{shi2020communication} was proposed to obtain an optimal merged size considering both gradient sparsification and overlapping between communication and computation. To obtain the optimal size of merged tensors, a mathematical model was analyzed and defined by verifying whether merging the neighbor layer shortens the training time. As a result, OMGS-SGD has the advantage of adaptively defining merging layers in various models. 

Many methods have been proposed to merge multiple layers into one large layer to effectively communicate with other processes in distributed deep learning. However, only a few methods have been proposed for gossip-based methods. In gossip-based methods, model segmentation is mainly used for selecting different network topologies on each layer/segment for direct diffusion of gradients/parameters. GossipGrad \cite{daily2018gossipgrad} was the first method proposed that considers the layer-wise selection of a partner (i.e., peer-to-send weight parameter) at each layer. In their study, Layer-wise selection of partners is considered in asynchronous SGD (AGD) \cite{dean2012large} which is comparing targets for GossipGraD; however, layer-wise selection is not applied in GossipGraD method. Instead, GossipGraD applies batch-wise gossip that selects a changed partner in each training step and changes the network topology at every $log_2(the\,number\,of\,worker\,nodes)$ train step to select diverse partner processes for a fast diffusion of weight parameters. COMBO \cite{hu2019decentralized} was proposed to reduce network bandwidth utilization in each communication. COMBO selects the worker process using a random network topology with constraints to select all workers at least once as peer when constructing the entire DNN model for the next step. However, this constraint on worker selection does not balance the network overhead among workers at each training step.

\section{Crossover-SGD}

Our analysis of gossip methods in the large mini-batch problem conditions with fixed number of worker nodes (explained later in Section 4.2) shows that as the number of mini-batch increases, gossip-based methods show higher validation accuracy than AllReduce-SGD. Thus, increasing the speed of parameter propagation of gossip-based SGD becomes a crucial point to utilize the high accuracy of gossip-based SGD when a large mini-batch problem occurs. Crossover-SGD is a novel gossip communication method that utilizes (i) overlapped segment-wise communication and (ii) random network topology with the following features.

\begin{itemize}
    \item \textbf{Overlapped segment-wise communication} : In COMBO \cite{hu2019decentralized}, segment-wise communication is helpful in utilizing the limited network bandwidth. In addition, in GossipGraD \cite{daily2018gossipgrad}, layer-wise gossip can be utilized for the fast propagation of weight parameters. By analyzing these two approaches, we conclude that segment-wise communication can more quickly propagate the weight parameter in the distributed deep learning environment and efficiently use network bandwidth resources in the local area network.

    \item \textbf{Random network topology with fair peer selection} : In GossipGraD \cite{daily2018gossipgrad}, the network topology is changed at every  $log_2(the\,number\,of\,worker\,nodes)$ step for direct diffusion of the weight parameter. Inspired by this approach, we utilize random network topology for our communication methods. However, random network topology does not fit well in distributed deep learning owing to the possibility of skewed communication among nodes. Thus, we propose a novel random network topology that avoids skewed peer selection among nodes.

\end{itemize}

Crossover-SGD is similar to COMBO in terms of segment-wise communication. However, COMBO mainly considers the distributed deep learning environment with low network bandwidth and effectively utilizes the limited network bandwidth with segment-wise communication. Furthermore, it does not consider fair peer selection between worker nodes well; thus, an skewed peer selection can increase the entire training time for distributed deep learning in the same local area network; in addition, COMBO does not need to consider the large mini-batch problem because it only considers the distributed deep learning environment with edge devices. Thus, each client needs to set small mini-batch sizes which do not cause the large mini-batch problem. However, this small local-mini-batch size cannot adapt to a common GPU cluster owing to its low utilization of GPUs. As a result, COMBO \cite{hu2019decentralized} is difficult to utilize in distributed deep learning with a large mini-batch problem setting. GossipGraD \cite{daily2018gossipgrad} is proposed for the fast propagation of the weight parameter, which is similar to the purpose of Crossover-SGD. However, Crossover-SGD utilizes different network topology and is based on segment-wise communication, unlike GossipGraD, which is based on batch-wise gossip.

When Crossover-SGD is compared to SGP \cite{assran2019stochastic}, which is the state-of-the-art gossip communication method for distributed deep learning, we observe that SGP has network topology to select peer to be fair similar to Crossover-SGD. However, SGP uses model-wise communication and directed exponential network topology to balance the network communication overhead and fair communication between worker nodes. These two differences affect the weight parameter propagation speed among nodes because the entire model parameter is sent to the worker node with directed exponential network topology that can select only limited workers in the worker node group. Thus, the validation accuracy of Crossover-SGD is higher than that of SGP for a large mini-batch problem.

\begin{figure}[ht]
\centering
\includegraphics[width=0.75\linewidth]{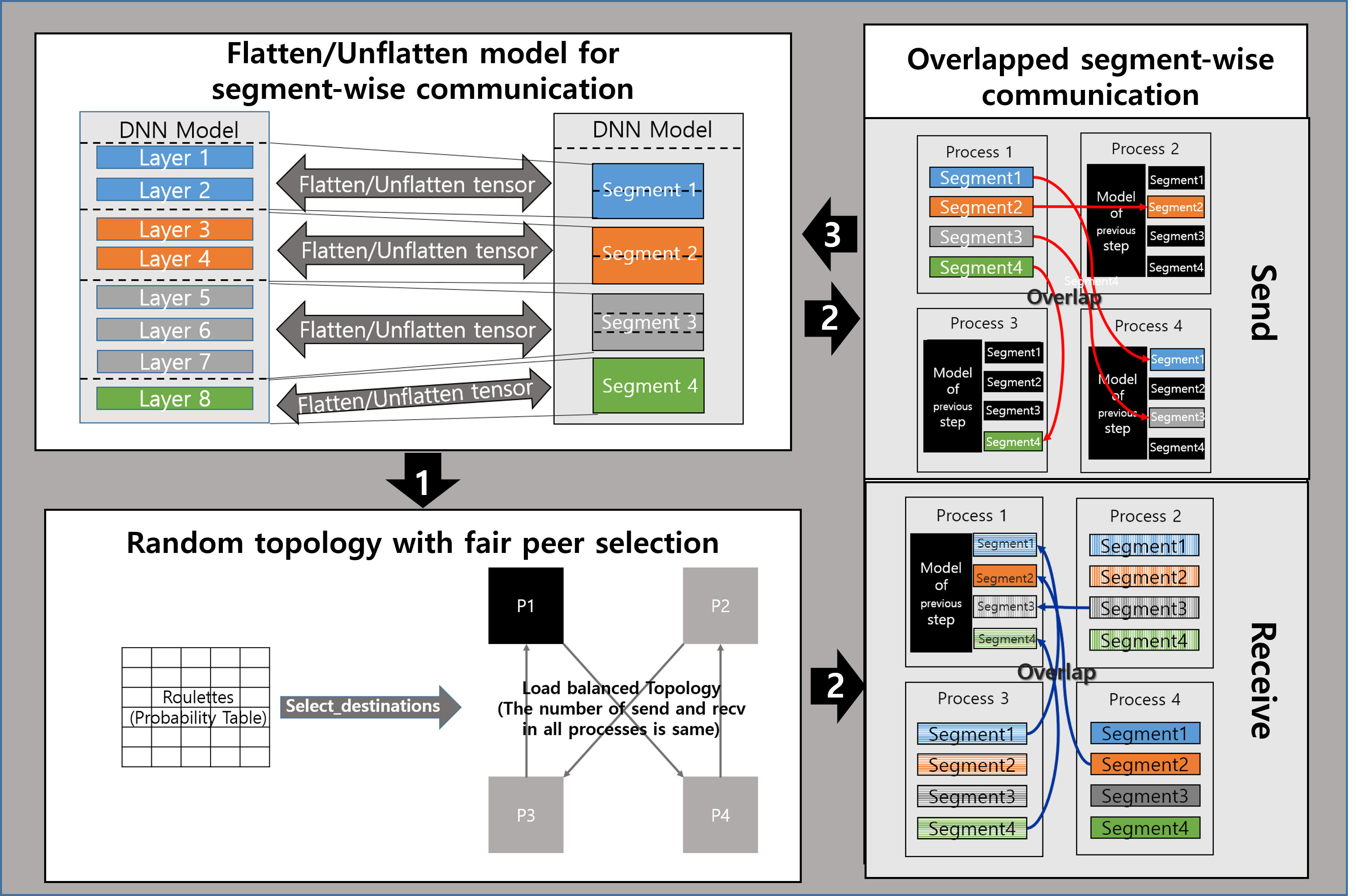}
\caption{Overview of Crossover-SGD\label{overview}}
\end{figure}

Crossover-SGD is a gossip communication method based on segment-wise communication and random network topology. Figure~\ref{overview} shows an overview of Crossover-SGD. We discuss the three parts of the Crossover-SGD in detail. The first part involves merging the series of layers into one large segment or recovering the tensors to apply the received tensor into a layer-wise model. In particular, a segment is a group of multiple layers in the model. Thus, we can more efficiently utilize the network bandwidth than layer-wise communication and reduce bottlenecks caused by model-wise communication.

The second part creates the random network topology for each segment. The flattened tensor (i.e., segment) is communicated among worker nodes based on created random network topology. If we set the same network topology on every segment of the model, the result of the communication is equal to that of model-wise communication. Thus, we set different random network topology for each segment, as shown in the bottom left of Figure~\ref{overview}

The third part is the overlapped segment-wise communication. After we set a random network topology for each segment, the worker nodes are communicated based on the random network topology. To use the network bandwidth of each node efficiently, we overlap the segment-wise communication. If the gossip communication process of the whole model is completed, the flattened tensor is flattened to multiple layers to merge the received segment to the DNN model of each rank. A detailed description of our proposed methods is presented in Sections 3.1 and 3.2. We also explain variants of our method utilizing hierarchical communication to increase node scalability with regard to validation accuracy in Section 3.3.

\subsection{Overlapped segment-wise communication of Crossover-SGD}

Segment-wise gossip communication \cite{hu2019decentralized} is an efficient method when the model suffers from network overhead. In segment-wise gossip communications, the series of layers are merged into one large tensor before communication. After communication of the merged tensor is finished, merged tensor is recovered to the original shape of the merged layers. However, it is difficult to overlap communication with computation because parameter averaging is processed after the gradient is applied to the weight parameters. Furthermore, it increases network latency compared to model-wise communication, which is widely used in distributed deep learning. This latency problem of segment-wise communication occurs because each layer communicates sequentially. For effective segment-wise communication, we overlap segment-wise communication using asynchronous send and receive.

\begin{algorithm}[ht]
\caption{Overlapped Segment-wise Communication}
\begin{algorithmic}[1]
\INPUT{ \textit{model segments}\,:\,segments\,of\,current\,training\,model,\,\textit{my rank}\,:\,rank\,ID\,of\,current\,process,\newline\textit{rseed}\,:\,random\,seed\, which\,is\,shared\,by\,every\,process,\,\textit{world\_size}\,:\,the\,total\,number\,of \, training\,process,\newline\textit{roulettes}\,:\,probability\,table\,of\,all\,processes\,which\,is\,used\,to\,select\,destination\,rank,\newline \textit{send\_list,\,recv\_list}\,:\,list\,for\,saving\,send\,and\,recv\,distributed\,request\,object }
\DESCRIPTION{ part\_model : partial DNN model corresponding to the current unflattened segment}  
\For{i=1...len(model,segments)} \Comment{request send and recv of each segment asynchronously}
\State $ segment \gets \textit{model\,segment[i]}$
\State $tensor\_flatten \gets flatten\_tensors(segment)$  \Comment{transform multiple tensors into one large tensor for communication}
\State $destinations = select\_destinations(rseed, world\_size, roulettes)$ \Comment{select destination rank of all ranks by random topology with fair peer selection}
\State $receive\_from = destinations[my\,rank]$
\State $send\_to = destinations.index[my\,rank]$
\State $completed\_send = isend(segment,send\_to)$ \Comment{asynchronously send segment}
\State $completed\_recv = irecv(segment,recv\_to)$ \Comment{asynchronously receive segment}
\State $append\,completed\_send\,to\,send\_list,\,append\,completed\_recv\,to\,recv\_list $

\EndFor
\State $n \gets 0$

\For { send,\, recv\,\, \textbf{in\,zip}\, (send\_list, recv\_list) }
\State $\textbf{wait}\,send\,and\,recv$ \Comment{wait until communication of all segments are completed}
\State $segment \gets \textit{model segment[n]}, n++;$
\State $tensor\_unflatten = unflatten\_tensors(segment)$ \Comment{transform one large tensor into multiple tensors for merging weight parameter to model}
\For{ layer, received\_layer\,\, \textbf{in\,zip}\, (part\_model, tensor\_unflatten) }
\State $ layer = (layer + received\_layer)\, / \, 2 $
\EndFor

\EndFor

\end{algorithmic}
\end{algorithm}

Algorithm 1 shows the process of overlapped segment-wise communication in Crossover-SGD. In lines 1–10, it sends and receives the segment of the model asynchronously. In line 13, it waits until all asynchronous peer-to-peer communication is completed before merging the received segments into the model. In this waiting step, it waits for send and recv as the order of requests of send and recv in lines 1-10. After all asynchronous communication is completed, the flattened tensor is transformed into unflattened tensors. Finally, these unflattened tensors are averaged using the weight parameter of the corresponding layer of each process model.

In proposed segment-wise communication, the network delay in the synchronization step (e.g., line 13 of Algorithm 1) can be a major bottleneck that delays training iterations. This synchronization delay can be alleviated by overlapping the communication delay with computations \cite{zhang2017poseidon}. In previous studies \cite{hashemi2019tictac, kang2020tensorexpress, shi2019mg}, efficient overlapping between the synchronization delay and computation is a difficult challenge due to the high network overhead caused by AllReduce-SGD. However, the synchronization step in our gossip communication can be easily overlapped with computations due to the network overhead caused by gossip communication does not increase linearly with the number of processes unlike AllReduce-SGD.

\begin{figure}[ht]
\centering
\includegraphics[width=0.6\linewidth]{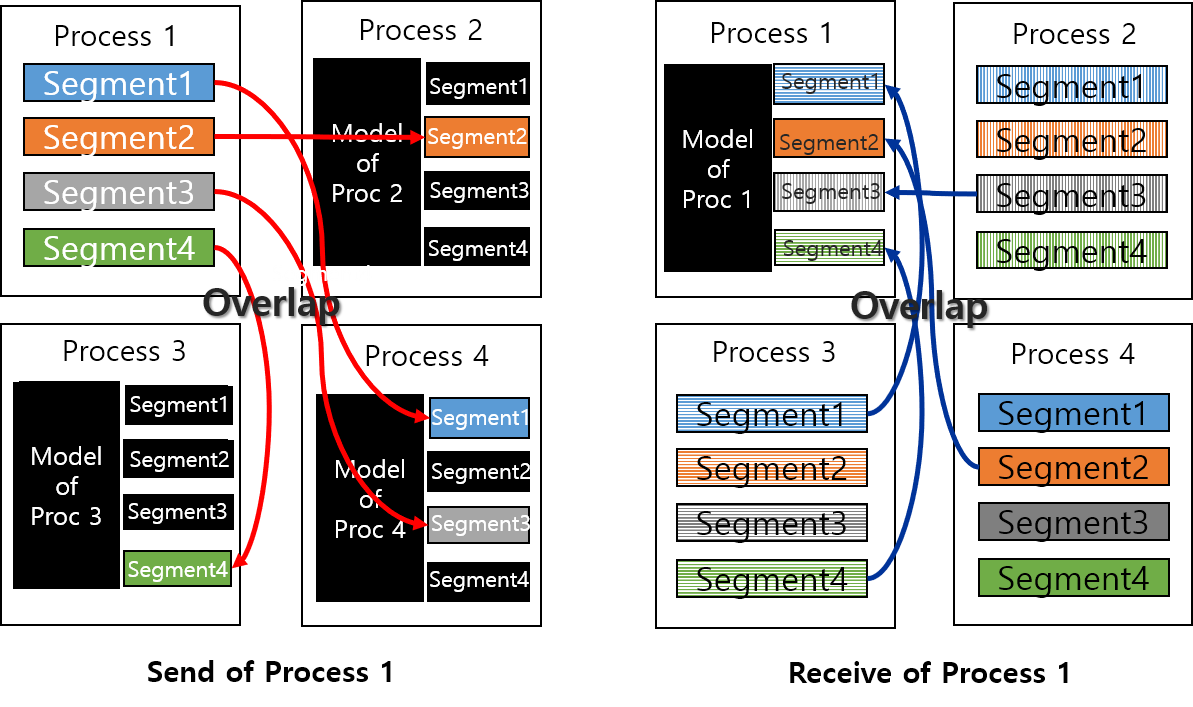}
\caption{Overview of segment-wise communication of Crossover-SGD\label{segcomm}}
\end{figure}

An overview of the segment-wise communication of Crossover-SGD is shown in Figure~\ref{segcomm}. In the figure, we represent the send and receive processes for each segment in terms of Process 1. In the sending process of the left part of the figure, each segment of Process 1 is randomly propagated into other processes. When the weight parameter of each segment is received by the destination process, the received weight parameter is merged with the weight parameter of Process 1, as shown in the right side of the figure. This segment-wise gossip communication can propagate the weight parameter more widely than model-wise communication as well as efficiently use of network bandwidth for each worker node.

\subsection{Random topology with fair peer selection}

\begin{algorithm}[ht]
\caption{Random topology with fair peer selection}
\begin{algorithmic}[1]
\INPUT{ \textit{rseed}\,:\,random\,seed\, which\,is\,shared\,by\,every\,process,\,\textit{world\_size}\,:\,the\,total\,number\,of \, training\,process,\newline\textit{roulettes}\,:\,probability\,table\,of\,all\,processes\,which\,is\,used\,to\,select\,destination\,rank,\,\newline\textit{dest\_list}\,:\,destination rank list of every worker node}
\DESCRIPTION{This algorithm is the selection\_destinations function in Algorithm 1}
\For{i...world\_size}
\For{j \textbf{in}\, dest\_list}
\State$ roulettes[i][j] = 0$
\EndFor
\State roulette\_sum = sum(roulettes[i])
\State roulettes[i] = [r/roulette\_sum for r in roulettes[i]]
\State selected\_rank = choice(world\_size,\,roulette=roulettes[i],\,seed=rseed)
\State dest\_list.append(selected\_rank)
\EndFor
\State return dest\_list
\end{algorithmic}
\end{algorithm}

Crossover-SGD allows one process to communicate with other processes on random network topology with fair peer selection. The topology is an essential part of the method that differentiates our proposed Crossover-SGD from COMBO and GossipGraD \cite{hu2019decentralized,daily2018gossipgrad}. To achieve this, we build a probability table, called roulettes, to manage the selection of peer worker nodes for communication. In roulettes, the row index represents the ID of the source process, and the column index represents the ID of the destination process. Furthermore, roulettes should avoid being selected by the peer workers. Therefore, the initial roulettes are created as a stochastic matrix with its major diagonal elements having zero value (i.e., zero possibility).

Even if roulettes (i.e., a shared probability table) are created, more procedures are required to build a network topology with fair peer selection. For example, a selected rank (i.e., process ID) should not be selected for another process again in one peer setting because duplicated selection means that some processes do not communicate in this training step. This unfair selection leads to a decrease in validation accuracy because uncommunicated workers that do not receive the weight parameters of other workers are less optimized than communicated workers in each time step.

\begin{figure}[ht]
\centering
\includegraphics[width=0.6\linewidth]{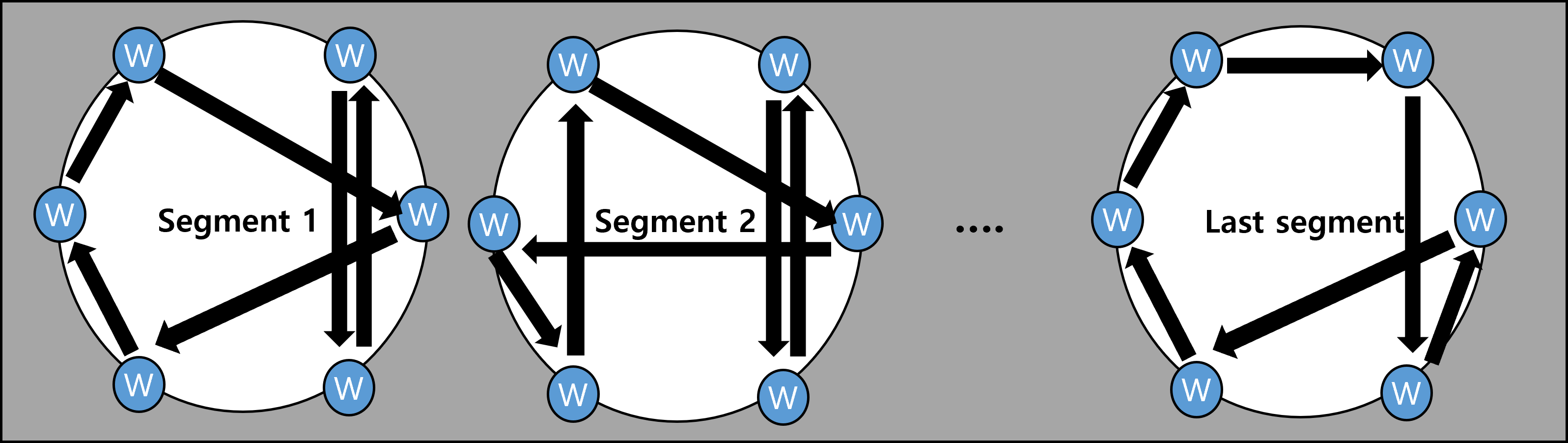}
\caption{Example of random network topology wit fair peer selection in Crossover-SGD\label{topo}}
\end{figure}

Algorithm 2 shows the process used to create a random network topology with fair peer selection. This algorithm corresponds to the select-destination function used in Algorithm 1. In lines 2 and 4, the process already selected as a destination has its probability set as zero to avoid duplicated selections. By the procedure between lines 2 and 4, the previously selected process cannot be chosen as the destination of another process. Thus, the segment of all processes is communicated at once with a randomly selected process and it can achieve a fair selection of peers. In lines 5 and 6, the roulettes for rank $i$ are reconfigured to reflect the removal of duplicated selections and the sum of probability is set as 1. After the roulettes for rank $i$ are fully modified, our algorithm selects the destination rank using reconfigured roulettes for rank $i$ in line 7. 
Using Algorithm 2, our random network topology can be defined as a doubly stochastic matrix whose convergence rate is $O(1/\sqrt{k})$ \cite{xi2017distributed, doan2020convergence}, 
whereas the conventional random network topology is a row-stochastic matrix whose convergence rate is $O(\log(k)/\sqrt{k})$\cite{zhang2019compressed,nedic2009distributed}. 
As a result, training with our random network topology converges faster than training with conventional random network topology.
Figure~\ref{topo} shows the example of the random network topology when the number of processes is six. As shown in figure, a new random network topology is created by each segment and each process receives only one segment. Furthermore, the destination of the segment in each process does not duplicate with other processes as explained in Algorithm 2. 

\begin{figure}[ht]
\centering
\includegraphics[width=0.7\linewidth]{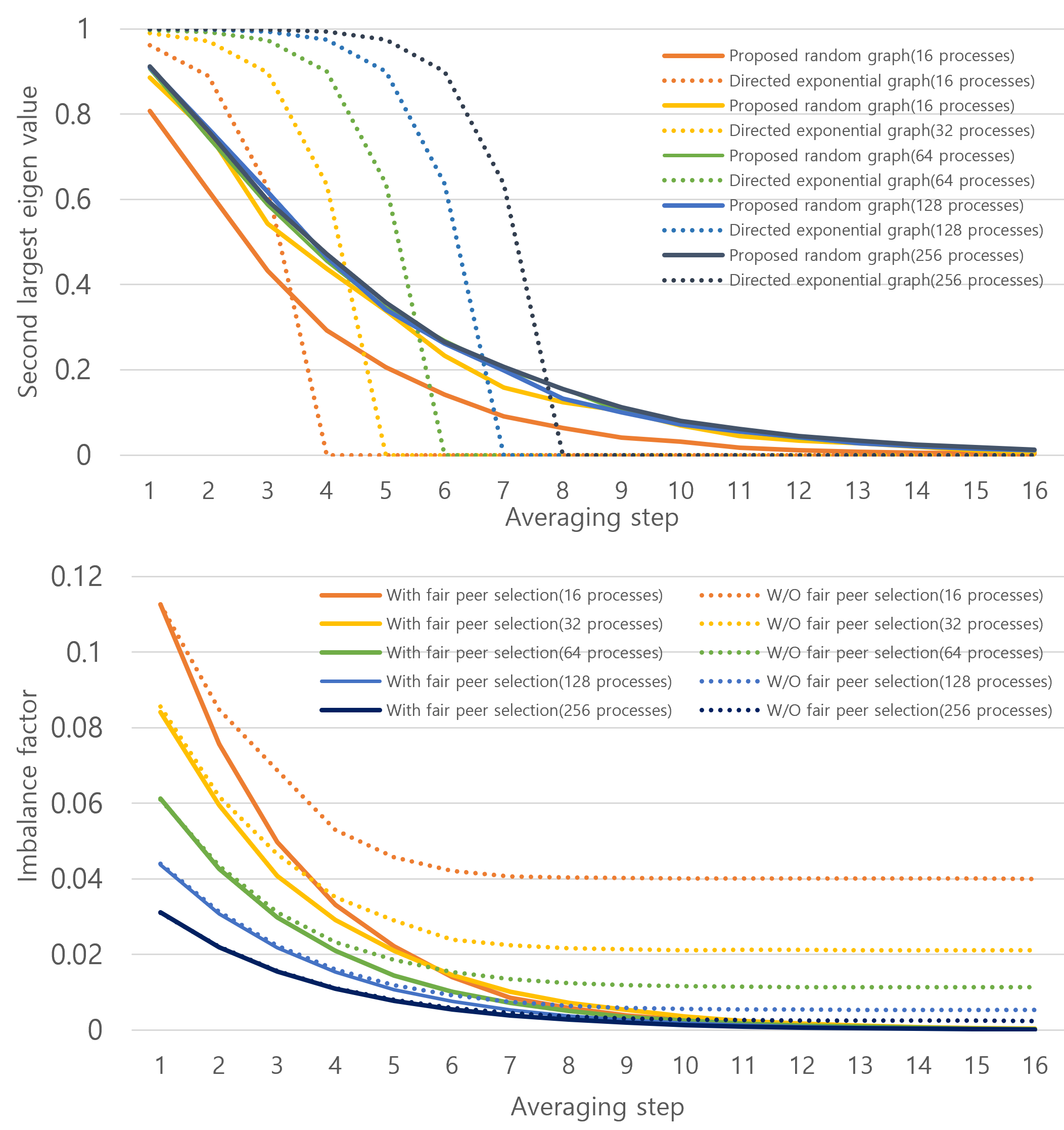}
\caption{Evaluation of proposed random network topology\label{randomtopo}}
\end{figure}

To compare averaging error of proposed random network topology with previous studies, we evaluate which is the upper bound of averaging error of mixing matrix in averaging step \cite{assran2019stochastic} using the second largest eigenvalue of $M^k$, the k th product of mixing matrix, $M$ and the evaluation results in the top graph in Figure~\ref{randomtopo} verified it. In this evaluation, we compare the $\lambda(2)$, the second largest eigen value of $M^k$, at each averaging step k for our proposed network topology and the directed exponential network topology. In the graph, we verified that our proposed topology has lower $\lambda(2)$ than the directed exponential network topology in the early steps. From the experiment result, we can see the directed exponential network topology suffers from a relatively higher averaging error than our proposed network topology at the early averaging steps. The higher averaging error can lead to the higher staleness (i.e., stale synchronous parallel). Since the staleness has the negative impact on the accuracy \cite{dai2015high, jiang2017heterogeneity, dai2018toward} (e.g., early step staleness yields exponential growth of the number of iterations for the same level of accuracy \cite{chen2016stochastic}), we can derive that the proposed network topology converges faster (i.e., performs better) than the directed exponential network topology because of its relative low averaging error.    

Furthermore, we analyze the imbalance of model averaging to evaluate the fair selection mechanism of our random network topology. We defined the imbalance factor as follows; When we defined the model parameters in the specific training step $i$ of process $j$ as $p_j^i$, we can define the model averaging results of $p_j^i$ after k step as $M^k{p_j^i}$. In addition, we can define the uniform matrix in which the sum of each row is one, $M_{AR}$ as the mixing matrix of AllReduce, which means that the model parameter is equally averaging with other processes. Because the information distribution via AllReduce can be defined as the objective of information averaging of gossip-based deep learning \cite{kong2021consensus, chen2021accelerating}, 
we can derive the imbalance between processes from the difference between $M^k$ and $M_{AR}$ as the imbalanced propagation among processes at each averaging step k. Interestingly, the imbalance factor of conventional random network topology does not converge to zero. This imbalance is proved by following Theorem 1 and 2.

\begin{lemm}
The convergence of the product of the random stochastic matrices results in a rank-one matrix. \cite{touri2013product} 
\end{lemm}
\begin{lemm}
The product of the stochastic matrices is a stochastic matrix. \cite{ANDRILLI2010491}
\end{lemm}

\begin{theorem}
The convergence result of the product of the random doubly stochastic matrices is equal to the AllReduce network topology.
\end{theorem}

\begin{proof}
Under Lemma 1 and Lemma 2, the product of a doubly stochastic matrices is converged to a doubly stochastic, rank-one matrix, which means that the value of all elements in the matrix is $1/N$ when $N$ is the number of processes. Thus, the convergence result is equal to the matrix of the AllReduce topology.
\end{proof}

\begin{theorem}
The convergence of the product of the row stochastic matrices results in identical rows which sum to one.
\end{theorem}
\begin{proof}
Under Lemma 1 and Lemma 2, the convergence of the product of row stochastic matrices results in a row stochastic, rank-one matrix. Rank-one matrix means that all rows in the matrix can be defined as the weighted value of row $i$, when $i$ is any row index of the matrix. Meanwhile, row stochastic means the sum of columns in any row $i$ is always one. However, a row stochastic, rank-one matrix does not guarantee the columns in the same row are equal.
\end{proof}

By Theorem 1, we can derive that the product of the proposed network topology is converged as the AllReduce topology. Meanwhile, the convergence of the product of conventional random network topology is not guaranteed to converge to zero by Theorem 2, because it is defined as the row-stochastic, rank-one matrix. Since each column in conventional random network topology is not guaranteed to sum to one. the model parameter in each process does not equally average with its peer in every iteration. This can be one of the causes of the conventional random network topology's propagation deviance.

In the bottom graph of Figure~\ref{randomtopo}, we define the variance value between these two matrices as an imbalance factor and verified that the imbalance of random network topology without peer fair selection is not converged with zero. As a result, we assumed that the random network topology with duplicated selection can decreased validation accuracy because this imbalance in training (e.g., imbalance dataset and heterogeneous computing performance between workers) need to be resolved in many previous studies \cite{chaudhary2020balancing, li2020federated, li2021federated}.

\subsection{Hierarchical communication methods for high validation accuracy}

We discussed our gossip-based SGD for the fast propagation of weight parameters in Sections 3.1 and 3.2. However, the validation accuracy is still decreased owing to the characteristics of gossip-based communication. The hierarchical communication method \cite{jia2018highly,kurth2017deep} was recently studied to increase scalability of the AllReduce-SGD by reducing the number of communications at each step. Likewise, we adapt a hierarchical communication method to Crossover-SGD to utilize our method in a large cluster environment. Interestingly, the network overhead is increased compared to the case of utilizing only the gossip-based method owing to the additional synchronization overhead in each group unlike in the cases of AllReduce-SGD. However, the validation accuracy can be preserved if hierarchical communication is used in gossip-based methods by adapting gossip communications to only inter-group communications.

\begin{figure}[ht]
\centering
\includegraphics[width=0.5\linewidth]{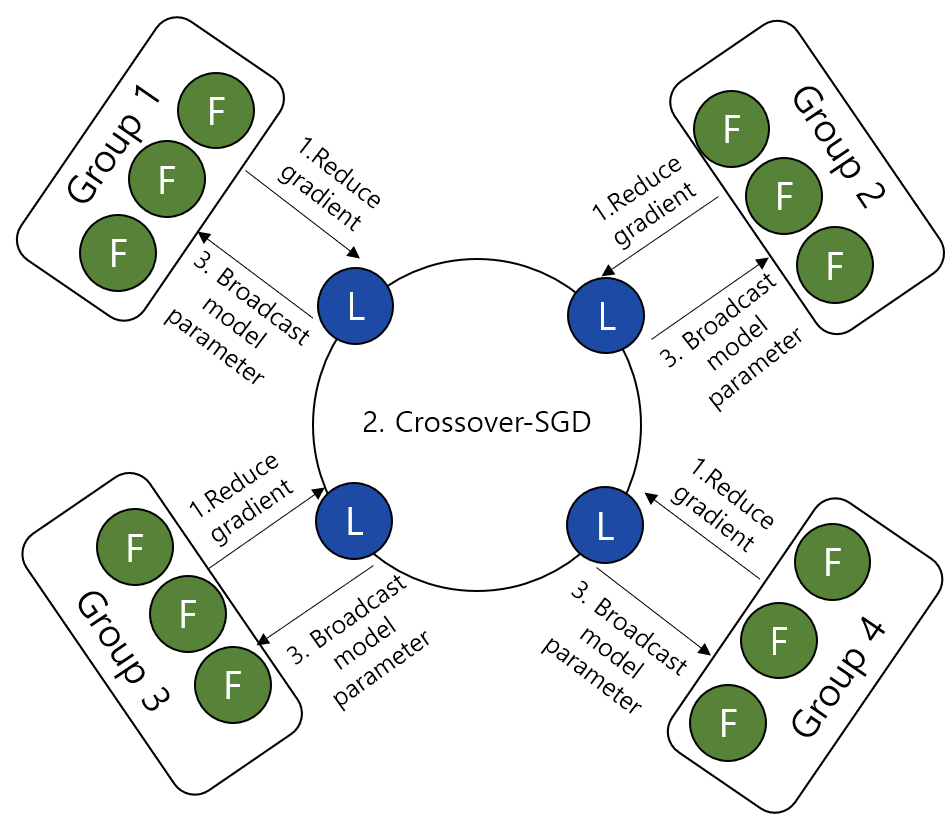}
\caption{hierarchical Crossover-SGD for node scalability about preserving high validation accuracy \label{group}}
\end{figure}

Figure~\ref{group} is the overview of hierarchical Crossover-SGD. In this figure, hierarchical Crossover-SGD is presented in three steps. The first step is to reduce the gradients of the worker nodes in each group. In this step, we reduced the gradients for two reasons. First, it utilizes computation and communication overlap, which can be utilized easily when gradients are communicated. Second, LARS \cite{you2017large}, when it is used with AllReduce-SGD, decreases the validation accuracy when the gradient norm is not synchronized among worker nodes. Thus, the L node, which is the leader node in each group, collects gradients from each worker and reduces them to apply a reduced gradient to the model. The second step is the communication of the inter-group by utilizing Crossover-SGD. Subsequently, the gossiped model parameter of the leader node propagates to other workers in each group.

\section{Evaluation}

In this section, we present the empirical experiment results in two fold because our evaluation has two goals: 1) to study the issue of the gossip-based approach in a large mini-batch environment, and 2) to measure the scalability and accuracy enhancement of our proposed method.

To achieve the first goal, we conducted an analysis on the performance of the gossip-based approach in a large mini-batch environment. The most intuitive way to perform the analysis is to set the condition as that of the large mini-batch problem. The ResNet-50 model \cite{he2016deep} with the ImageNet ILSVRC 2012 dataset \cite{russakovsky2015imagenet} is commonly used as the base model and dataset. Reproducing the large mini-batch problem is difficult to configure for the ResNet50 model owing to the economic cost of utilizing a large GPU cluster. Thus, we chose to increase the communication interval and accumulate the loss during this interval instead of increasing the physical scale. The increase in communication interval has almost the same effect as a large mini-batch environment except for the statistics update of batch normalization that incurs a decreasing validation accuracy when the mini-batch size of each worker becomes smaller. 

To verify that our proposed Crossover-SGD effectively addresses the large mini-batch problem, which is our second goal, we conduct empirical comparison experiments between our method and conventional schemes such as AllReduce-SGD and SGP. The AllReduce-SGD scheme is the most representative method for distributed learning, and the gradient of every model is synchronized at each training interval. Thus, the accuracy of this approach is known to be the most stable compared to any other communication approach. SGP \cite{assran2019stochastic} is a state-of-the-art gossip-based communication scheme for distributed learning. The training time of this approach is known to be the fastest, however, its validation accuracy decreases when the number of worker nodes increases.

\subsection{Evaluation Environment}

In our evaluation, we ensure that the starting point of the valuation is the state-of-the-art research result of a large mini-batch problem. Most recently, Fujitsu achieved approximately 75.9\% validation in 86,016 mini-batch sizes. Thus, we first set the validation accuracy of our ResNet-50 model with AllReduce equal to or slightly lower than the convergence accuracy. As a result, we achieved a 75.67\% accuracy in our model with AllReduce-SGD. To achieve 75.67\% accuracy, the implementation of LARS is critical to the performance. To increase accuracy, we implemented the LARS of MXNet on PyTorch. This implementation showed the highest accuracy compared to any other user implementation of LARS on PyTorch.

A summary of the hyperparameter settings is shown in Table~\ref{hyper}. For our evaluation, we used the ResNet-50 model with the ImageNet ILSVRC 2012 dataset. Our hyperparameter setting was based on the hyperparameter setting of Fujitsu \cite{fujitsu}. As shown in Table~\ref{hyper}, we set the communication interval as 42 to simulate the condition of a large mini-batch problem in a small cluster environment because the maximum mini-batch size in V100 is limited to hundreds with a GPU memory size 16 or 32 GB. In terms of data augmentation of input data, we only utilized random resizing, random cropping, and horizontal flipping for our training. These augmentations were originally adapted from the official example code of Resnet-50 in PyTorch \cite{paszke2019pytorch}. In addition, we used mixed precision training \cite{jia2018highly} to reduce the training time.

\begin{table}[ht]
\centering
\caption{Hyperparameter setting in Training with 86,016 mini-batch size}
\begin{tabular}{ | c | c | }
\hline
Hyper-parameter Type & Value\\ \hline
LARS Coefficient & 0.0025\\ \hline
Learning rate & 9\\ \hline 
Optimizer & Adam with $5\times10^{-5}$ weight decay and 0.96 momentum \\ \hline
Warm-up epoch & 36 \\ \hline
training epoch & 90 \\ \hline
Communication Interval & 42 \\ \hline
Mini-batch size per Node & 256\\ \hline
Data augmentation & Random resize crop , Horizontal flip \\ \hline
Model segment & Segmenting blocks and FC layer of ResNet \\ \hline
Mixed precision training & Yes \\ \hline
$\epsilon$ in Label smoothing & 0.1 \\ \hline
\end{tabular}
\label{hyper}
\centering
\end{table}

The evaluation is executed on the NEURON GPU cluster environment of KISTI \cite{neuron} as shown in Table~\ref{spec}. Because we utilize only four worker nodes in this cluster environment, we create multiple processes for each GPU device in the evaluation discussed in Section 4.3. For evaluating time efficiency, we utilize the AWS cluster environment because the fact that the impact of network overhead is relatively small in small cluster environment of NEURON with InfiniBand. The worker nodes in the AWS cluster environment are communicated by low network bandwidth and GPUs, which have a relatively lower performance than V100 GPU. For the evaluation, we use PyTorch 1.4, with Apex \cite{apex} as the distributed deep learning framework and OpenMPI 3.1.0 \cite{openmpi} as the communication library. To utilize the cluster environment, we use the container library, singularity 3.1.0 \cite{kurtzer2017singularity}. We use the ResNet-50 model in torchvision \cite{torchvision}, which is an official DNN model and dataset library of PyTorch.

\begin{table}[ht]
\centering
\caption{Hardware Specification}
\begin{tabular}{ | c | c |}
\hline

\multicolumn{2}{ | c | }{\textbf{Hardware for Section 4.2 and 4.3 except for the evaluation in Figure ~\ref{time}}} \\ \hline 
CPU &  Intel Xeon Cascade Lake (Gold 6226R) / 2.90GHz (16-core) / 2 socket \\ \hline
GPU & NVIDIA Tesla V100 32GB X 2 \\ \hline
RAM & DDR4 384GB \\ \hline
Network Bandwidth & 56000Mb/s \\ \hline
\multicolumn{2}{ | c | }{\textbf{Hardware for the evaluation in Figure ~\ref{time} (i.e., AWS g4dn.2xlarge \cite{g4} EC2 Instance)}} \\ \hline 
CPU &  AWS-custom Second Generation Intel® Xeon® Scalable (Cascade Lake) processors(8-core) \\ \hline
GPU & NVIDIA Tesla T4 \\ \hline
RAM & 32 GiB \\ \hline
Network Bandwidth & up to 3125Mb/s \\ \hline
\end{tabular}
\label{spec}
\centering
\end{table}

\subsection{Analysis of gossip methods in the large mini-batch problem condition}

\begin{figure}[ht]
\centering
\includegraphics[width=1.0\linewidth]{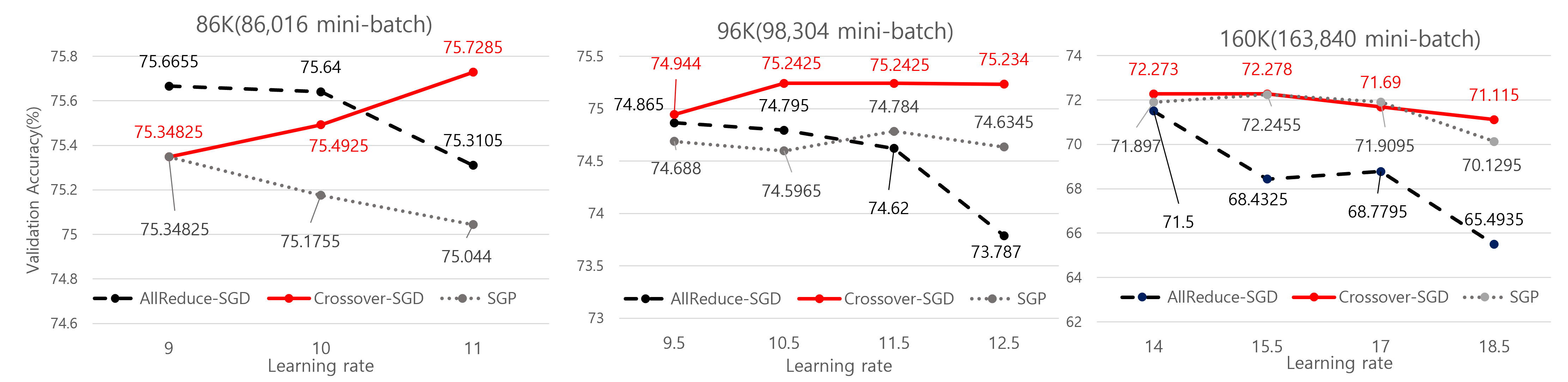}
\caption{ Impact of learning rate when the mini-batch size is increased\label{learningrate}}
\end{figure}

To analyze the gossip methods, we compare two gossip methods with our baseline model. First, we determined the optimal learning rate of each global mini-batch size. Generally, this optimal learning rate increases linearly with increasing global mini-batch size. Figure~\ref{learningrate} shows a comparison of each method. In this figure, we can derive and confirm two important insights. First, a linear increase in the learning rate does not effectively work in a large mini-batch setting. In our evaluation with increasing mini-batch size, we did not increase the learning rate linearly with increasing mini-batch size; however, we increase the learning at a much smaller rate than in the linear increasing case. Second, each method has a different optimal learning rate. The second insight shows that each method has different validation accuracies for each learning rate. Most importantly, our proposed method, Crossover-SGD, shows the most robust results on changes of learning rate, whereas AllReduce-SGD shows the most fluctuating result on learning rate changes.

\begin{figure}[ht]
\centering
\includegraphics[width=1.0\linewidth]{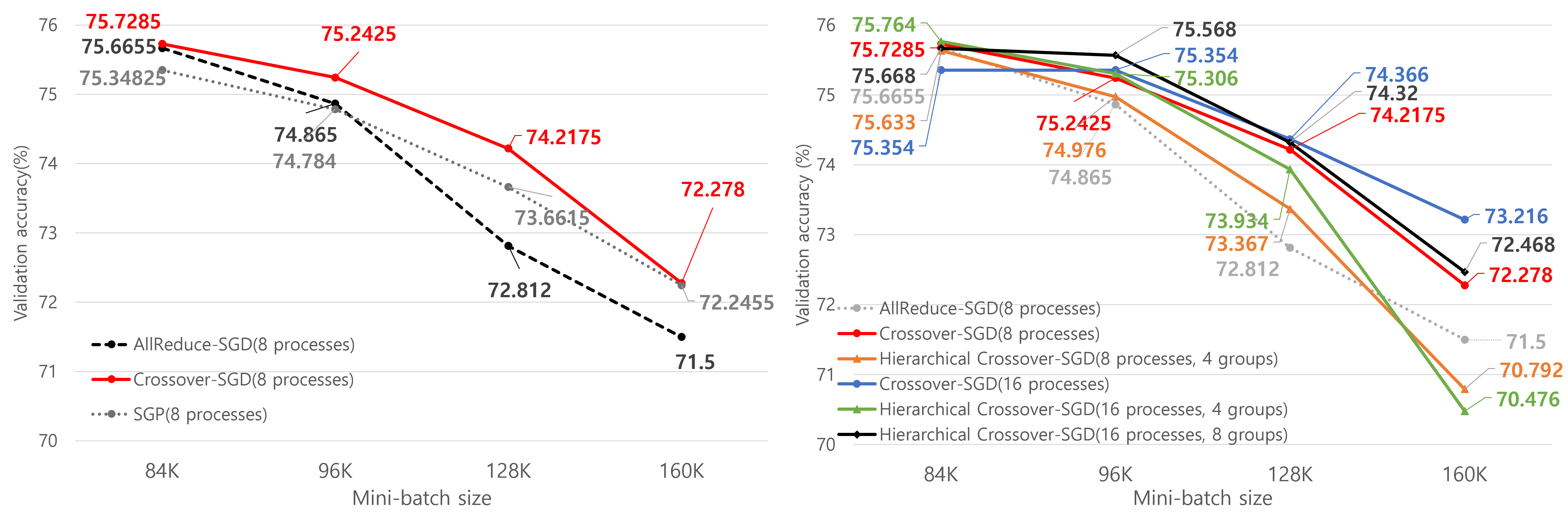}
\caption{Validation accuracy with increasing of mini-batch size for the large mini-batch problem\label{increasing}}
\end{figure}

Figure~\ref{increasing} compares the accuracy of the methods for the mini-batch size with the optimal learning rate of each method. Crossover-SGD showed the highest accuracy compared to the other methods. As shown in Figure~\ref{increasing}, AllReduce-SGD was most affected by the large mini-batch problem. Conversely, two gossip-based methods, Crossover-SGD and SGP are more robust than AllReduce-SGD in terms of the large mini-batch problem, if the number of worker nodes is fixed. This is an interesting result because we assumed that AllReduce is the most stable communication method. However, we confirmed that the exploration characteristic of the gossip-based method is helpful in preserving validation accuracy. As the mini-batch size is increased, the difference in validation accuracy with Crossover-SGD and SGP decreases. This is because the random network topology results in unstable convergence of the neural network when the number of training steps is decreased. In summary, the two gossip methods show higher validation accuracy than AllReduce-SGD when a large mini-batch is utilized with a limited number of workers. This means that gossip methods can be utilized for large mini-batch conditions if the node scalability problem is alleviated.

When comparing hierarchical Crossover-SGD with other methods, the converged validation accuracy of hierarchical Crossover-SGD is closer to AllReduce-SGD as it has a lower number of groups. However, the hierarchical Crossover-SGD with 4 groups shows the steep decrease in validation accuracy when the mini-batch size is 160k. This means that when the number of groups is reduced, hierarchical Crossover-SGD is unstable as its large mini-batch problem is severe. On the other hand, when group size is relatively large (e.g., eight group size), this sharp decrease does not show. In hierarchical Crossover-SGD, the number of groups means that the number of the participating processes in gossip communications. It assumes that it has a high possibility to record the same validation accuracy when the number of groups in hierarchical Crossover-SGD is equal to other hierarchical Crossover-SGD cases or the number of processes in Crossover-SGD. This assumption is verified in right graph of Figure~\ref{increasing} when it compares orange and green lines or red and black lines.

\subsection{Node scalability comparison in the large mini-batch problem}

\begin{figure}[ht]
\centering
\includegraphics[width=1.0\linewidth]{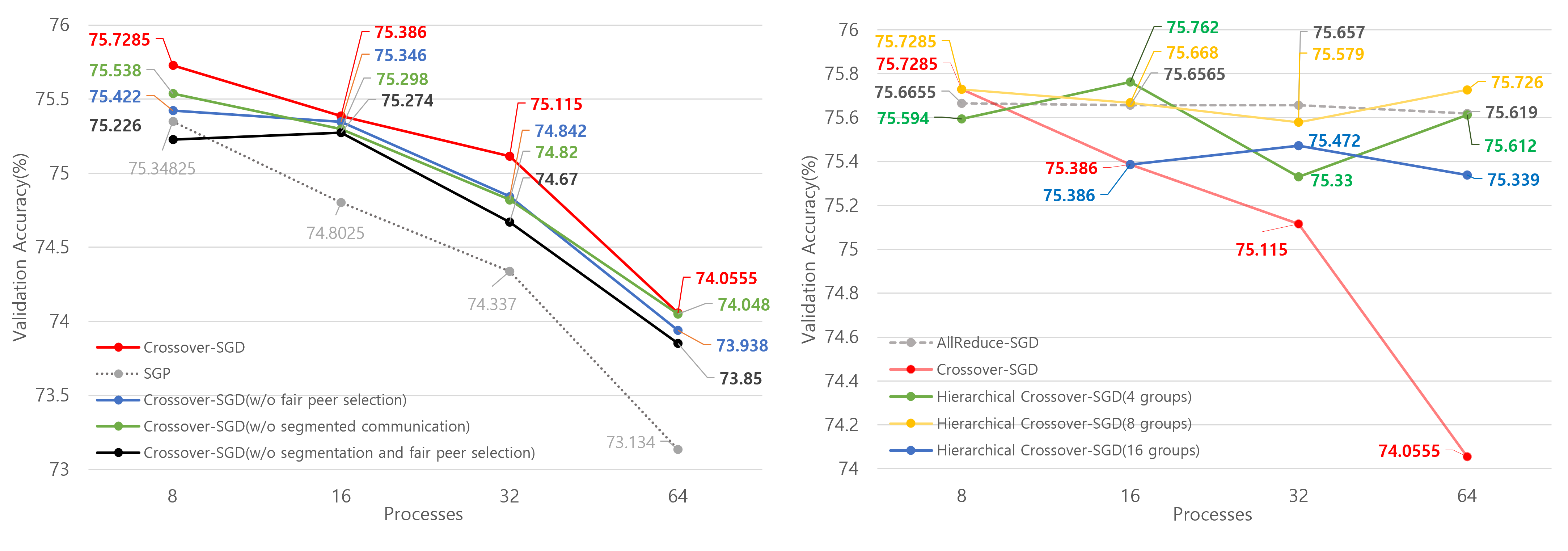}
\caption{Validation accuracy with increasing number of worker processes and fixed mini-batch size (i.e., 86,016) \label{nodes}}
\end{figure}

In this section, we evaluate the node scalability of Crossover-SGD in both time and validation accuracy. First, Figure~\ref{nodes} shows the comparison of validation accuracy when the number of worker processes is increased, and the mini-batch size is fixed. As the number of worker processes is increased, the validation accuracy of two gossip-based methods, Crossover-SGD and SGP, decreases. This is due to the delayed propagation of weight parameters of the gossip-based algorithm that worsens as the number of worker processes increases. However, Crossover-SGD is more robust with an increase in the number of nodes compared to the SGP. We assumed that this is because of the fast propagation of weight parameters caused by the two characteristics of our methods: segment-wise communication and random network topology.

As we explained in Section 3.2, Crossover-SGD without fair peer selection or segmented-wise selection shows reduced validation accuracy in every number of processes. Furthermore, we verified the interesting results when comparing Crossover-SGD without segment-wise communication and fair peer selection (e.g., the case which only adapts random network topology) with SGP (e.g., directed exponential graph). In this comparison, we verified that the random network topology shows higher validation accuracy than the directed exponential graph when the number of peers is equal. On the other hand, the random network topology shows the lowest validation accuracy when compared with other variations of Crossover-SGD in the left graph of Figure~\ref{nodes}. It means that fair peer selection and segment-wise communication can be used together.

Crossover-SGD is more robust when the number of nodes is increased. However, it is apparent that the validation accuracy of the gossip methods also decreases. We assume that this perceptible decrease in the validation accuracy originates from the delayed propagation of the model parameters by the gossip methods. To alleviate this problem, we applied hierarchical grouping to Crossover-SGD. The right graph of Figure~\ref{nodes} shows the comparison between hierarchical Crossover-SGD with other comparing groups. In this graph, we verified that our hierarchical Crossover-SGD has similar validation accuracy to Crossover-SGD if the number of groups is equal to the number of processes in Crossover-SGD. In detail, Hierarchical Crossover-SGD with 16 groups (i.e., the blue line in the right graph of Figure~\ref{nodes}) shows similar accuracy with Crossover-SGD with 16 processes, and this tendency is maintained in other Hierarchical Crossover-SGD cases as well. 
In the case of 64 processes, there is a sudden drop in validation accuracy in both Crossover-SGD and SGP. This phenomenon can be easily found in related works \cite{assran2019stochastic,lian2017can,kong2021consensus,koloskova2019decentralized,hegedHus2021decentralized}. 
However, it should be noted that the performance enhancement is distinctive in comparison to SGP and our proposed Crossover-SGD as we increase the number of processes that participate in the model training.

\begin{figure}[ht]
\centering
\includegraphics[width=1.0\linewidth]{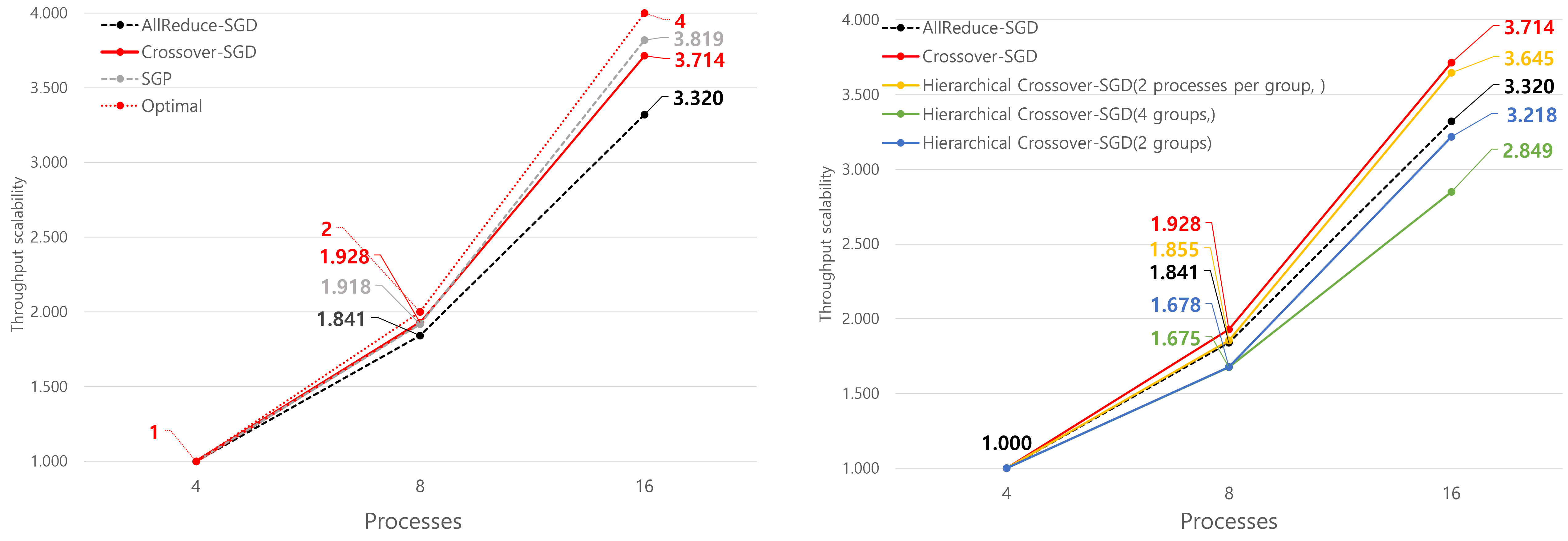}
\caption{Time efficiency comparison about training time (model : ResNet-50)\label{time}}
\end{figure}

Figure~\ref{time} shows the comparison of time efficiency among the evaluated methods. In this evaluation, the optimal line shows the optimal time efficiency cases when the number of nodes is increased. As can be seen in this figure, the two gossip-based methods, Crossover-SGD and SGP, show approximately 12-14\% higher time efficiency than AllReduce. The two gossip methods show higher scalability than the AllReduce method owing to the low network overhead of the gossip-based method. However, Crossover-SGD has approximately 1-2\% lower scalability than SGP. This decrease is due to the different computation overhead to select segment-wise peer. As a result, Crossover-SGD and SGP shows linear scalability to the number of worker nodes. 

In the cases of hierarchical Crossover-SGD is in the right graph of Figure~\ref{time}, the scalability of hierarchical Crossover-SGD is lower than AllReduce-SGD when the number of groups remains the same. This is because the network overhead in group communications (e.g., broadcast and reduce) is linearly increased when the number of groups remains the same and the entire number of processes is increased. On the other hand, hierarchical Crossover-SGD shows higher scalability than AllReduce-SGD when the number of processes in a group remains the same. This means that increased inter-group communications (e.g., Crossover-SGD) cause less communication overhead compared to when the network overhead in the group is increased. Thus, the hierarchical Crossover-SGD should be optimized by improved group-wise network topologies such as 2d-torus topology \cite{mikami2018massively, geng2018hips, lin2021delivering} or be considered in heterogeneous network environment such as the network bandwidth between groups is low \cite{hong2018decentralized, wu2021deep, cho2019blueconnect}.

\section{Conclusion}

In this study, we analyzed the characteristics of gossip-based SGD on large mini-batch problems. We observed that gossip-based SGD achieves higher validation accuracy owing to its exploration characteristics when the number of workers is fixed. Thus, the delay propagation of gossip-based SGD becomes crucial when using gossip-based methods for large cluster environments. Based on these insights, we proposed Crossover-SGD, a gossip-based communication method that utilizes both the random network topology with fair peer selection and overlapped segment communication. These two characteristics of Crossover-SGD make it possible to propagate the weight parameter to other workers more efficiently and faster than previous gossip-based methods. We also conducted empirical evaluations and found that Crossover-SGD outperforms AllReduce-SGD in large mini-batch problems; it also shows higher accuracy than the state-of-the-art gossip-based method SGP for distributed deep learning when both the number of workers and local mini-batch size is increased.

 However, there is still scope for improvement in the scalability of Crossover-SGD. In terms of time efficiency, we can overlap the asynchronous communication of Crossover-SGD with the backward propagations. By overlapping between communications and computations, we can reduce the impact of synchronization step caused by waiting for asynchronous communications. However, it needs to schedule the communications to fulfill its dependency with computations. In terms of delaying the propagation of weight parameters. Even if it can achieve faster propagation than SGP, owing to the delay propagation of weight parameters in gossip-based methods, it is observed that the validation accuracy decreases significantly when the number of workers is increased. Thus, we apply hierarchical grouping communication for worker processes to alleviate the decrease in validation accuracy of gossip-based methods. Crossover-SGD with this grouping method (i.e., hierarchical Crossover-SGD) shows similar validation accuracy as AllReduce-SGD. Therefore, Hierarchical Crossover-SGD is a better alternative communication method to AllReduce-SGD, reducing network overhead efficiently while preserving high validation accuracy. Our code is publicly available in GitHub https://github.com/Yeosangho/Crossover-SGD

\section*{Acknowledgements}
This research was supported by National Supercomputing Center with supercomputing resources including technical support (KSC-2019-CRE-0105), the MSIT (Ministry of Science and ICT), Korea, under the ITRC (Information Technology Research Center) support program (IITP-2020-2018-0-01431) supervised by the IITP (Institute for Information \& communications Technology Promotion), Basic Science Research Program Through the National Research Foundation of Korea (NRF) funded by the Ministry of Education (2018R1D1A1B07043858), and Electronics and Telecommunications Research Institute (ETRI) grant funded by the Korean government (20ZT1100, Development of ICT Convergence Technology based on Urban Area). We would like to thank Editage (www.editage.co.kr) for English language editing.

\nocite{*}
\bibliography{ada/CCPECrossover.bib}%

\end{document}